\documentclass{ecai}
\usepackage{times}
\usepackage{graphicx}
\usepackage{latexsym}

\usepackage{amsmath}
\usepackage{amssymb}

\usepackage{amsmath,amssymb}
\DeclareMathOperator{\E}{\mathbb{E}}
\usepackage{subfigure}
\usepackage{mathtools}
\usepackage{booktabs} 
\usepackage{adjustbox}
\usepackage{algorithmic}
\usepackage{algorithm2e}
\ecaisubmission   

\begin{document}

\title{Dropout-GAN: Learning from a Dynamic Ensemble of Discriminators}

\author{Gon\c{c}alo Mordido\institute{Hasso Plattner Institute, Germany, email: goncalo.mordido@hpi.de} \and Haojin Yang\institute{Alibaba Group, China, email: haojin.yhj@alibaba-inc.com} \and Christoph Meinel\institute{Hasso Plattner Institute, Germany, email: christoph.meinel@hpi.de}}

\maketitle
\bibliographystyle{ecai}

\begin{abstract}
We propose to incorporate adversarial dropout in generative multi-adversarial networks by omitting or dropping out the feedback of each discriminator with some probability at the end of each batch. Our approach forces the generator not to constrain its output to satisfy a single discriminator, but, instead, to satisfy a dynamic ensemble of discriminators. We show that the proposed framework, named Dropout-GAN, leads to a more generalized generator, promoting variety in the generated samples and avoiding the mode collapse problem commonly experienced with generative adversarial networks (GAN). We provide evidence that applying adversarial dropout promotes sample diversity on multiple datasets of varied sizes, mitigating mode collapse on several GAN approaches.
\end{abstract}

\section{Introduction}
\label{introduction}
Generative adversarial networks \cite{GANs}, or GAN, is a framework that integrates adversarial training in the generative modeling process. According to its original proposal~ \cite{GANs}, the framework is composed of two models - one generator and one discriminator - that train together by playing a minimax game. While the generator tries to fool the discriminator by producing fake samples that look realistic, the discriminator tries to distinguish between real and fake samples better over time, making it harder to be fooled by the generator.

However, one of the main problems with GAN is mode collapse~\cite{cross_domain_gans,towards,mode_regularized_gans,wgan}, where the generator is able to fool the discriminator by only producing data coming from the same data mode, \textit{i.e.,} connected components of the data manifold.
This leads to a poor generator that is only able to produce samples within a narrow scope of the data space, resulting in the generation of only similarly looking samples.
Hence, at the end of training, the generator comes short regarding learning the full data distribution, and, instead, is only able to learn a small segment of it.
This is the main issue we try to tackle in this work.


In a disparate line of work, dropout was introduced by \cite{dropout} and it has been proven to be a very useful and widely used technique in neural networks to prevent overfitting~\cite{understanding_dropout,warde2013empirical,gal2016theoretically}. In practice, it simply consists of omitting or dropping out, the output of some randomly chosen neurons with a probability $d$ or dropout rate. The intuition behind this process is to ensure that neurons are not entirely dependent on a specific set of other neurons to produce their outputs. Instead, with dropout, each neuron relies on the population behavior of several other neurons, promoting generalization in the network. Hence, the overall network becomes more flexible and less prone to overfitting.

The main idea of this work consists of applying the same dropout principles to generative multi-adversarial networks. This is accomplished by taking advantage of multiple adversarial training, where the generator's output is dependent on the feedback given by a specific set of discriminators. 
By applying dropout on the feedback of each discriminator, we force the generator to not rely on a specific discriminator or discriminator ensemble to learn how to produce realistic samples. Thus, the generator guides its learning from the varied feedback given by a dynamic ensemble of discriminators that changes at every batch.

In our use case, one can then see mode collapse as a consequence of overfitting to the feedback of a single discriminator, or even a static ensemble of discriminators. Hence, by dynamically changing the adversarial ensemble at every batch, the generator is stimulated to induce variety in its output to increase the chances of fooling the different possible discriminators that may remain in the ensemble at the end.
Our main contributions can be stated as follows:
\begin{itemize}

\item We propose a novel and generic framework, named Dropout-GAN (Section~\ref{dropout_gan}), that trains a single generator against a dynamically changing ensemble of discriminators. 
\item We provide useful discussions and insights regarding the benefits of multiple adversarial training in GAN, namely the increase training stability (Section~\ref{details}).
\item We test our method on several datasets and multiple metrics, showing that it succeeds in reducing mode collapse by promoting sample diversity within epochs (Sections~\ref{results} and \ref{metric_evaluation}).
\item We show that proposed approach of applying adversarial dropout also improves several other GAN approaches on several metrics and datasets of different size and nature (Section~\ref{sec:comparison}), confirming the extensibility of our framework.
\end{itemize}


\section{Generative Adversarial Networks}
\label{gans}

As originally proposed~\cite{GANs}, the standard GAN framework consists of two different models: a generator ($G$), that tries to capture the real data distribution to generate fake samples that look realistic, and a discriminator ($D$), that tries to do a better job at distinguishing real and fake samples. $G$ maps a latent space to the data space by receiving noise as input and applying transformations to it to generate unseen samples, while $D$ maps a given sample to a probability $p$ of it coming from the real data distribution.

In the ideal setting, given enough iterations, $G$ would eventually start producing samples that look so realistic that $D$ would not be able to distinguish between real and fake samples anymore. Hence, $D$ would assign $p = 0.5$ to all samples, reaching a full state of confusion. However, due to training instability, this equilibrium is hard to reach in practice.
The two models play the following minimax game:

\begin{equation}
\label{gans_minimax}
\begin{split}
\min_{G}\max_{D}V(D,G) = \E_{x \sim p_{r}(x)}[\log D(x)] \\ + \E_{z \sim p_{z}(z)}[\log (1-D(G(z)))],
\end{split}
\end{equation}

where $p_{z}(z)$ represents the noise distribution used to sample $G$'s input and $G(z)$ represents its output, which can be considered as a fake sample originated from mapping the modified input noise to the data space. On the other hand, $p_{r}(x)$ represents the real data distribution and $D(x)$ represents the output of $D$, \textit{i.e.,} the probability $p$ of sample $x$ being a real sample from the training set.

In order to maximize Eq.~\ref{gans_minimax}, $D$'s goal is then to maximize the probability of correctly classifying a sample as real or fake by getting better at distinguishing such cases by assigning $p$ close to 1 to real images and $p$ close to 0 to generated images. By contrast, to minimize Eq.~\ref{gans_minimax}, $G$ tries to minimize the probability of its generated samples being considered as fake by $D$, through fooling $D$ into assigning them a $p$ value close to 1.

However, in practice $\log (1-D(G(z)))$ might saturate due vanishing gradient problems in the beginning of training caused by $D$ being able to easily distinguish between real and fake samples. As a workaround, the authors propose to maximize $\log (D(G(z)))$ instead, making it no longer a minimax game.
Nevertheless, $G$ still continues to exploit $D$'s weaknesses in distinguishing real and fake samples by using $D$'s feedback to update its parameters and slightly change its output to more likely trick $D$ in the next iterations. 

\section{Dropout-GAN}
\label{dropout_gan}

\begin{figure}[h]
\centering
 \subfigure[Original GAN]{\includegraphics[scale=0.25]{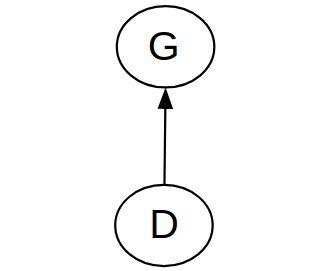}}\quad
  \subfigure[Dropout-GAN]{\includegraphics[scale=0.25]{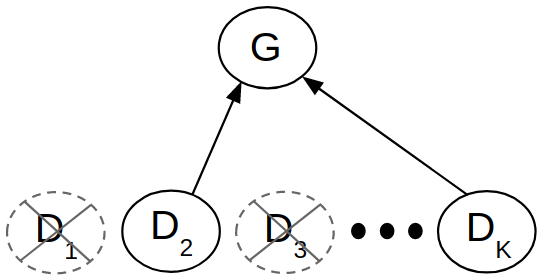}}
\caption{We expand the original GAN framework (left) to multiple adversaries, where some discriminators are dropped out according to some probability (right), leading to only a random subset of feedback (represented by the arrows) being used by $G$ at the end of each batch.}
\label{fig:dropout_gan}
\end{figure}

We propose to integrate adversarial feedback dropout in generative multi-adversarial networks, forcing $G$ to appease and learn from a dynamic ensemble of discriminators. This ultimately encourages $G$ to produce samples from a variety of modes, since it now needs to fool the different possible discriminators that may remain in the ensemble. Variations in the ensemble are achieved by dropping out the feedback of each $D$ with a certain probability $d$ at the end of every batch. This means that $G$ will only consider the loss of the remaining discriminators in the ensemble while updating its parameters at each iteration. Figure~\ref{fig:dropout_gan} illustrates the proposed framework.

Our initial modification to the value function $V$ of the minimax game is presented in equation (\ref{eq:gans_minimax_proposed_1}), where $\delta_k$ is a Bernoulli variable ($\delta_k \sim Bern(1-d)$) and $\small\{D_k\small\}$ is the set of $K$ total discriminators. The gradients calculated from the loss of a given discriminator $D_k$, are only used for the calculation of $G$'s final gradient updates when $\delta_k = 1$, with $P(\delta_k = 1) = 1-d$. Otherwise, this information is discarded:

\begin{equation}
\begin{split}
\min_{G}\max_{\big\{D_k\big\}} \sum_{k=1}^{K} V(D_k,G) =\sum_{k=1}^{K} \delta_k(\E_{x \sim p_{r}(x)}[\log D_k(x)]\\+ \E_{z \sim p_{z}(z)}[\log (1 - D_k(G(z)))]).
\end{split}
\label{eq:gans_minimax_proposed_1}
\end{equation}

There is, however, the possibility of all discriminators being dropped out from the set, leaving $G$ without any guidance on how to further update its parameters. In this case, we randomly pick one discriminator $D_j \in \big\{D_k\big\}$ and follow the original objective function presented in equation (\ref{gans_minimax}), using solely the gradient updates related to $D_j$'s loss to update $G$.
Hence, taking into consideration this special case, our final value function, $F$, is set as follows:

\begin{equation}
\begin{split}
F(G, \big\{D_k\big\}) =\\
    =\begin{dcases*}
              \min_{G}\max_{\big\{D_k\big\}} \sum_{k=1}^{K} V(D_k,G) ,\quad \text{if }  \exists k: \delta_k = 1\\
              \min_{G}\max_{D_j}V(D_j,G) ,\quad\text{otherwise}, \text{for } j \in \big\{1, ..., k\big\}.
            \end{dcases*}
\end{split}
\label{gans_minimax_proposed_2}
\end{equation}

It is important to note that each discriminator trains independently, \textit{i.e.,} is not aware of the existence of the other discriminators, since no changes were made on their individual gradient updates. This implies that even if dropped out, each $D$ updates its parameters at the end of every batch. The detailed algorithm of the proposed solution can be found in Algorithm \ref{alg:dropout_gan}.

\begin{algorithm}
\caption{Dropout-GAN. }
\begin{algorithmic}
\STATE {\bfseries Initialize:} $m \leftarrow \frac{B}{K}$

\FOR{\textbf{each} iteration}

\FOR{$k = 1$ to $k = K$}
\STATE {} \textbullet~Sample minibatch $z_i$, $i=1 \ldots m$, $z_i\sim p_g(z)$
\STATE {} \textbullet~Sample minibatch $x_i$, $i=1 \ldots m$, $x_i\sim p_{r}(x)$


\STATE {} \textbullet~Update $D_k$ by ascending along its gradient:
$$\nabla_{\theta_{D_k}} \frac{1}{m} \sum_{i=1}^{m} [\log D_k(x_i) +\log (1-D_k(G(z_i)))]$$

\ENDFOR

\STATE {} \textbullet~Sample minibatch $\delta_k$, $k=1 \ldots K$, $\delta_k\sim Bern(1-d)$



\IF{$all(\delta_k) = 0$}

\STATE {} \textbullet~Sample minibatch $z_i$, $i=1 \ldots m$, $z_i\sim p_g(z)$

\STATE {} \textbullet~Update $G$ by descending along its gradient \quad from a random discriminator $D_j$, for some $j \in \big\{1, ..., K\big\}$:
$$\nabla_{\theta_G} \frac{1}{m} \sum_{i=1}^{m} \log (1-D_j(G(z_i)))$$
\ELSE 

\STATE {} \textbullet~Sample minibatch $z_{k_i}$, $i=1 \ldots m$,\quad\quad $k=1 \dots K$, $z_{k_i}\sim p_g(z)$
    
\STATE {} \textbullet~Update $G$ by descending along its gradient:
$$\nabla_{\theta_G} \sum_{k=1}^{K} \delta_k(\frac{1}{m} \sum_{i=1}^{m} \log (1 - D_k(G(z_{k_i}))))$$
\ENDIF
\ENDFOR
\end{algorithmic}
\label{alg:dropout_gan}
\end{algorithm}

\section{Implementation Details}
\label{details}

In this section, we provide a detailed study of the effects of using a different number of discriminators together with different dropout rates. Moreover, we further provide insights into the consequence of splitting the batch among the different discriminators on the generator's training. 
The proposed framework was implemented using Tensorflow~\cite{tensorflow2015-whitepaper}.



\subsection{Number of Discriminators}

Training instability has been noticeably reported as one of GAN biggest problems. Here, we show that this problem can be eased by using multiple adversaries. This is also stated in previous works~\cite{gman,multiple_random_projection_gans}, however, without much detailed evidence. Furthermore, on top of increasing training stability, using multiple discriminators enables the usage of the original $G$ loss, since there is now an increased chance that $G$ receives positive feedback from at least one $D$ and is able to guide its learning successfully~\cite{gman}.

To analyze the training procedure, we correlate the degree of training instability with the gradient updates that are being used by $G$ to update its parameters at the end of each batch. The intuition is that if such updates are big, the parameters of the model will change drastically at each iteration. This is intuitively an alarming sign that the training is not being efficient, especially if it still occurs after several epochs of training, since $G$ is repeatedly greatly updating its output, instead of performing slight, mild changes in a controlled fashion.

We found that when using multiple discriminators such gradients would converge to zero as training progressed, while, on the contrary, they remained high (in terms of its absolute value) when using solely one discriminator. On the other hand, we also noticed that as the number of discriminators increases, the point at which $G$'s gradients start to converge also increases, suggesting that using more discriminators can delay the learning process. However, this is expected since $G$ now receives more (and possibly contradictory) feedback regarding its generated samples, needing more time to utilize such information wisely.

\subsection{Batch Partitioning}

The main purpose of splitting the batch among the different discriminators is to encourage each to specialize in different data modes.  This is achieved by training them with a different subset of samples of the same size within each batch. This applies to both the fake samples produce by $G$ and real samples retrieved from the training set. Such partitioning also allows data parallelism, diminishing the overhead caused by using more discriminators in the framework.

To further investigate the success in forcing the different discriminators to focus on different data modes, we argue that $G$'s capacity of fooling the ensemble should decrease in such situation. This is indeed confirmed in our experiments, with $G$'s loss being higher when the batches are split, especially later on in training where each $D$ had enough time to focus on a single or a small subset of data modes. Thus, one can then associate the higher $G$ loss with the generated samples now having to comply with a higher number of realistic features to be able to fool the dynamic ensemble of discriminators, with a subset of such features being used by each $D$ to characterize a given sample as real or fake.



We increase the overall batch size to enable each $D$ to be trained on the same original number of samples at each batch.
On the other hand, $G$ might still have access to all samples at each batch, since it uses the feedback from the remaining discriminators to update its parameters at the end. 
However, having weaker discriminators by training each one of them with fewer samples than $G$ is not necessarily bad since they are more likely to give positive feedback to $G$~\cite{gman,multiple_random_projection_gans}. This is a result of their possible confused state that can better aid $G$ in producing realistic samples than if it would continuously receive negative feedback, especially in the long run.

\subsection{Dropout Rate}

Dropping out the loss of a given $D$ with a probability $d$ before updating $G$'s parameters is what induces variability in our framework. This forces $G$ not to only need to fool one or even a static set of discriminators, but, instead, to fool a dynamic ensemble of adversaries that changes at every batch. Hence, performing this type of dropout can also be seen as a form of regularization, since it aims to promote more generalizability on the fake samples produced by G.

Depending on the number of discriminators used, using a small probability $d$ of dropout might only lead to small changes in the ensemble of adversaries, making the feedback seen by $G$ nearly constant throughout every batch. On the other hand, using a large dropout probability might lead to too much variance in the ensemble, making it difficult for $G$ to learn properly due to the variability of the visible set. 

Evidence of the correlation between the dropout rate and the quality of the generated samples is further given in Sections~\ref{results} and \ref{metric_evaluation}.
Similarly to what was discussed in the original Dropout paper~\cite{dropout}, we found that using $d = 0.2$ and $d = 0.5$ often led to better results, both in a qualitative and quantitative manner. Nevertheless, we also found that using any dropout rate ($0 < d \leq 1$) consistently performed better across the different datasets than when using a static ensemble of adversaries ($d = 0$). 

\section{Experimental Results}
\label{results}

We tested the effects of the different parameter settings on three different datasets: MNIST~\cite{mnist}, CIFAR-10~\cite{cifar10}, and CelebA~\cite{celeba}. We compared all possible combinations by using the different number of discriminators across the set $\small\{1, 2, 5, 10\small\}$ with each different dropout rate in $\small\{0.0, 0.2, 0.5, 0.8, 1.0\small\}$. We used the DCGAN inspired architecture used in GMAN \cite{gman}, with $G$ consisting of 4 convolutional layers of decreasing neuron size, \textit{e.g.,} 128, 64, 32, 1 (for MNIST) or 3 (for CIFAR-10 and CelebA), and each $D$ having 3 convolutional layers of increasing number of neurons, \textit{e.g.,} 32, 64, 128, and a fully connected layer at the end. We refer to GMAN \cite{gman} for more information regarding the training settings. Important to note that, even though all discriminators share the same architectures, their weights are initialized differently. Results are reported below for each dataset.

\subsection{MNIST}
\label{mnist}

\begin{figure}[h]
\centering
\includegraphics[width=0.45\textwidth]{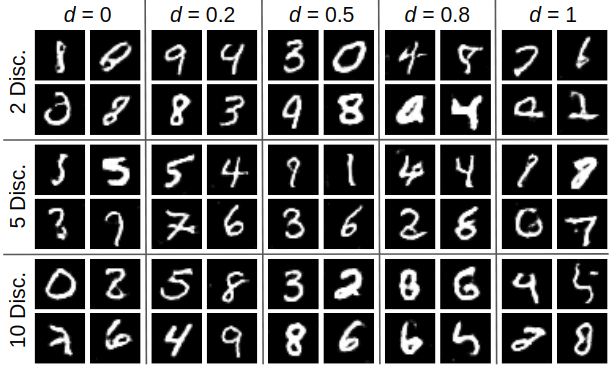}
\caption{MNIST results using different combinations of the number of discriminators and dropout rates.}
\label{fig:mnist_results_summary}
\end{figure}

MNIST is composed of 10 different classes of handwritten digits varying from 0 to 9, with the generated samples of Dropout-GAN being shown in Figure~\ref{fig:mnist_results_summary}.

It is visible that the quality and variation of the produced samples increase while using dropout rate values of 0.2 and 0.5 across all different sized discriminator sets. On the other hand, the quality of the produced numbers deteriorates considerably while using high dropout rates, \textit{i.e.,} 0.8 and 1, or no dropout rate at all. However, the quality gets slightly better when using more discriminators on such extreme end dropout rates, since $G$ might still get enough feedback to be able to learn at the end of each batch.

\subsection{CIFAR-10}
\label{cifar10}

\begin{figure}[h]
\centering
\includegraphics[width=0.45\textwidth]{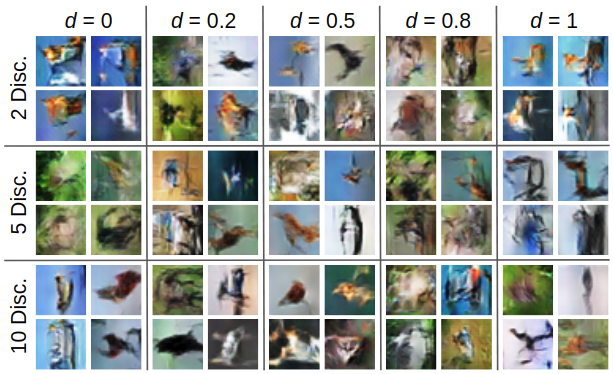}
\caption{CIFAR-10 results using different combinations of the number of discriminators and dropout rates.}
\label{fig:cifar_results}
\end{figure}

To further validate our solution, we used the CIFAR-10 dataset also composed of 10 classes, consisting of different transportation vehicles and animals. Results are presented in Figure~\ref{fig:cifar_results}. Once again, we observe worst sample quality when using high or nonexistent dropout values. Moreover, there are also clear traits of mode collapsing while using no dropout rate throughout all numbers of discriminators in the set. Sharper and more diverse samples are obtained while using 0.2 or 0.5 dropout rate and a bigger number of discriminators in the set.

\subsection{CelebA}
\label{celebA}

\begin{figure}[h]
\centering
\includegraphics[width=0.45\textwidth]{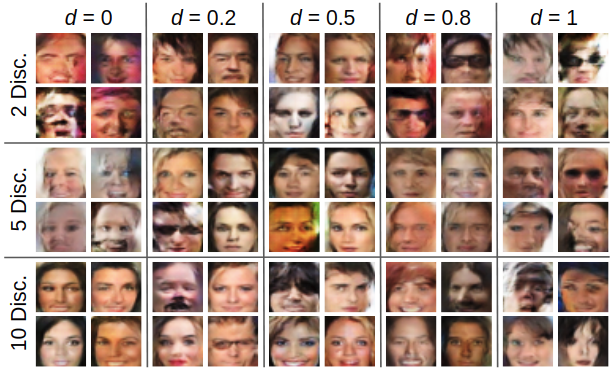}
\caption{CelebA results using different combinations of the number of discriminators and dropout rates.}
\label{fig:celeba}
\end{figure}

We lastly tested our approach in the cropped version of CelebA, containing faces of real-world celebrities. Results are given in Figure~\ref{fig:celeba}. One can see that using no dropout rate leads to similar looking faces, especially when using 2 and 5 discriminators. Once more, faces produced with mid-ranged dropout values with bigger discriminator ensembles present more variety and sample quality than the rest.

\section{Parameter Evaluation}
\label{metric_evaluation}

Since the results shown above rely heavily on subjective judgment, we now evaluate the effects of using a different number of discriminators and dropout rates on each dataset in a quantitative way. Note that the presented results are not state-of-the-art since exploring several architectural settings is not the focus of this work. Instead, by using different architectures on different datasets, our focus is to compare the effect of the different parameter combinations.

\subsection{Fr\'echet Inception Distance}
\label{fid}

We used the Fr\'echet Inception Distance~\cite{fid} (FID) to measure the similarity between the fake and real images. The returned distance uses the mean $\mu$ and covariance $cov$ of a multi-variate Gaussian produced from the embeddings of the last pooling layer of the Inception-v3 model ~\cite{inception} for both the real data and the generated data. In the original paper, the authors show that FID is more robust to noise and more correlated to human judgment than Inception Score~\cite{is}. Moreover, FID has shown to be sensitive to mode collapse~\cite{are_all_created_gans_equal}, with the returned distances increasing when samples from certain classes are missing from the generated set.


\bigbreak
\textbf{Minimum FID.}
Table~\ref{min_table} shows the minimum FID obtained by $G$ for each dataset. Lower values indicate more similarity between the fake and real data. We ran all of our experiments for 40 epochs in total and used the same architecture described previously. To obtain the best FID across all epochs, we generated $10000$ samples from $G$ at the end of each epoch and then proceeded to calculate the FID between the set of the generated samples per epoch and the whole training set.

\begin{table}
\caption{Minimum FID obtained across 40 epochs using the different datasets. Bold scores represent the minimum FID obtained for each dataset. Underlined scores indicate the best FID within using a given number of discriminators and the different possible dropout rates regarding each dataset.}
\begin{center}
\begin{small}
\begin{sc}
\centering
\begin{adjustbox}{width=0.47\textwidth}
\begin{tabular}{lcccr}
\toprule
 & MNIST & CIFAR-10 & CelebA\\
\midrule
1 disc.    &  21.71 $\pm$ 0.39& 104.19 $\pm$ 0.07 & 53.38 $\pm$ 0.03\\
\midrule
2 disc.; $d=0.0$ & 24.88 $\pm$ 0.13& 106.54 $\pm$ 0.38 & 52.46 $\pm$ 0.08\\
2 disc.; $d=0.2$ & 22.34 $\pm$ 0.29& 103.55 $\pm$ 0.13 & 46.60 $\pm$ 0.03\\
2 disc.; $d=0.5$ & 22.08 $\pm$ 0.09&  \underline{103.20 $\pm$ 0.05} & \underline{45.90 $\pm$ 0.04}\\
2 disc.; $d=0.8$ & \underline{21.87 $\pm$ 0.10}& 103.60 $\pm$ 0.03 & 46.82 $\pm$ 0.14\\
2 disc.; $d=1.0$ & 23.56 $\pm$ 0.29& 104.73 $\pm$ 0.19 & 51.17  $\pm$ 0.01\\
\midrule
5 disc.; $d=0.0$ & 21.47 $\pm$ 0.40& 95.75 $\pm$ 0.15 & 45.89 $\pm$ 0.05\\
5 disc.; $d=0.2$ & 21.70 $\pm$ 0.12& 90.59 $\pm$ 0.35 & \underline{\textbf{36.36 $\pm$ 0.11}}\\
5 disc.; $d=0.5$ & \underline{19.25 $\pm$ 0.12}& \underline{89.74 $\pm$ 0.35} & 38.10 $\pm$ 0.54\\
5 disc.; $d=0.8$ & 20.26 $\pm$ 0.07& 90.77 $\pm$ 0.70 & 41.22 $\pm$ 0.24\\
5 disc.; $d=1.0$ & 20.54 $\pm$ 0.15& 95.71 $\pm$ 0.03 & 41.56 $\pm$ 0.18\\
\midrule
10 disc.; $d=0.0$ & 22.62 $\pm$ 0.10& 99.91 $\pm$ 0.10 & 43.85 $\pm$ 0.30\\
10 disc.; $d=0.2$ & 19.12 $\pm$ 0.01& 91.31 $\pm$ 0.16 & 41.74 $\pm$ 0.14\\
10 disc.; $d=0.5$ & \underline{\textbf{18.18 $\pm$ 0.44}}& \underline{\textbf{88.60 $\pm$ 0.08}} & \underline{40.67 $\pm$ 0.56}\\
10 disc.; $d=0.8$ & 19.33 $\pm$ 0.18& 88.76 $\pm$ 0.16 & 41.74 $\pm$ 0.03\\
10 disc.; $d=1.0$ & 19.82 $\pm$ 0.06& 93.66 $\pm$ 0.21 & 41.16 $\pm$ 0.55\\
\bottomrule
\end{tabular}
\end{adjustbox}
\end{sc}
\end{small}
\end{center}
\label{min_table}
\end{table}

By analyzing Table~\ref{min_table}, we observe that the minimum values of FID for all datasets were mostly obtained when using $d = 0.5$.
However, by analyzing the local minima obtained while maintaining the same number of discriminators and only varying the dropout rate, it is also noticeable that one can also generally achieve very competitive results while using $d \in \big\{0.2, 0.5, 0.8\big\}$, depending on the number of discriminators and datasets being used.
The results also show that applying dropout on multiple discriminators always leads to a better FID rather than maintaining the ensemble of discriminators static, \textit{i.e.} $d = 0$, or singular, \textit{i.e.,} using solely 1 discriminator.

\begin{figure*}[h]
\centering
\includegraphics[width=1.0\textwidth]{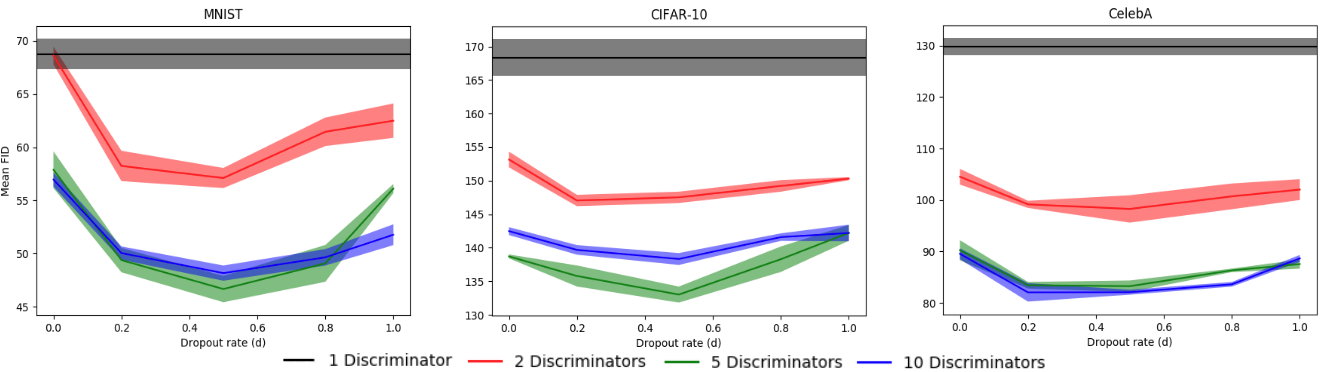}
\caption{Mean FID calculated across 40 epochs on the different datasets. Smaller values mean better looking and more varied generated samples over time. The convex representation of FID indicates the benefits in using mid-range dropout rates.}
\label{fig:mean_fid_results}
\end{figure*}

\bigbreak
\textbf{Mean FID.}
We followed the same procedure and calculated the mean FID across all 40 epochs. Results are presented in Figure~\ref{fig:mean_fid_results}. This evaluation promotes a broader look at the stage of $G$ at the end of every epoch, reflecting the quality and variety of the generated samples over time. The presented graphs provide a clear vision regarding the advantages of using multiple discriminators instead of solely one, with the FID being better in the first case. Using 5 or 10 discriminators with mid-range dropout rates leads to better FID results across all datasets. 

The similar looking performance when using 5 and 10 discriminators can be explained by what was previously mentioned regarding $G$ needing more time to learn from more feedback. Nevertheless, by analysis of Table~\ref{min_table} it is visible that better generated samples are produced when using 10 discriminators on all datasets, even if it takes more training to reach that state.
This ultimately means that by having access to more feedback, $G$ is eventually able to produce better, varied samples in a more consistent manner over time.

\begin{figure*}[h]
\centering
\includegraphics[width=1.0\textwidth]{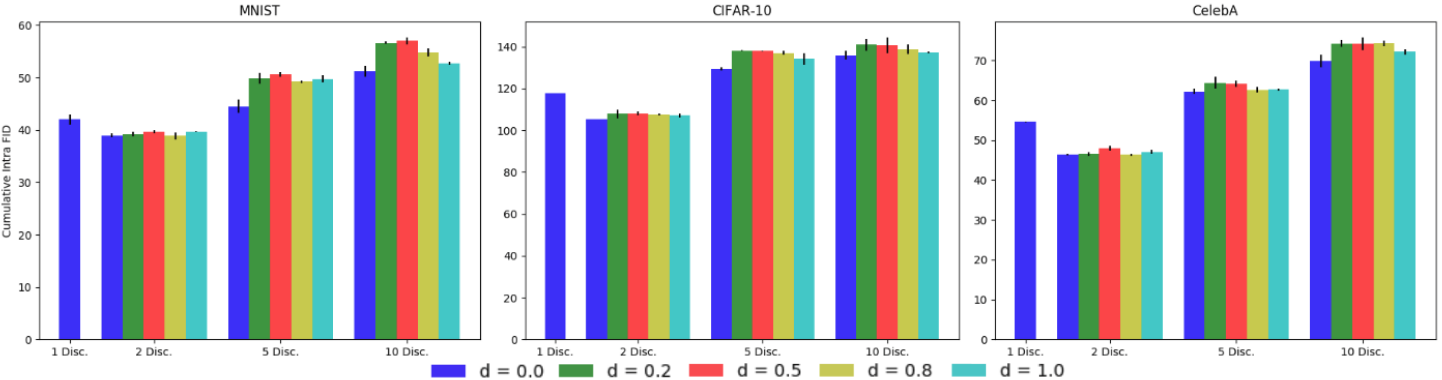}
\caption{Cumulative Intra FID across 40 epochs on the different datasets. Higher values represent more diversity over time.}
\label{fig:intra_cum_fid_results}
\end{figure*}

\bigbreak
\textbf{Cumulative Intra FID.}
To test the sample diversity within a given epoch, we calculated the FID between the set of generated samples of every epoch. This was accomplished by generating $20000$ samples from $G$ at the end of every epoch and then calculating the FID between the two halves of the generated set. We evaluated the diversity of the generated samples over time by adding all calculated FIDs for each model. Results are shown in Figure~\ref{fig:intra_cum_fid_results}.


From the analysis of the presented bar graphs, one can see the effect of using a different number of discriminators, with bigger sets of discriminators promoting a wider variety of generated samples within each epoch. This is generally observed across all datasets. Furthermore, it is noticeable the benefits of using mid-range dropout rates to promote sample diversity, especially when using a bigger discriminator set.

\section{Method evaluation}
\label{sec:comparison}

\begin{figure}[h]
  \subfigure[Toy dataset.]{\includegraphics[scale=0.3]{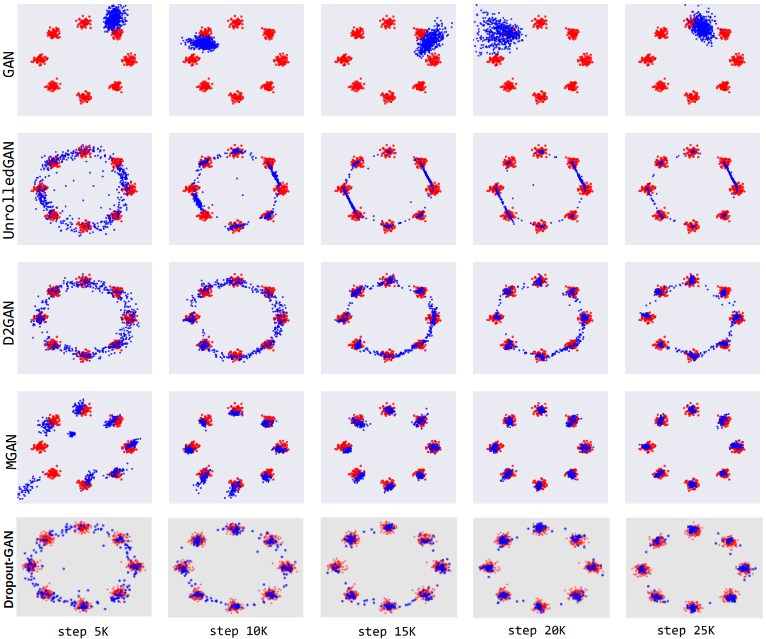}}
  \subfigure[Wasserstein distance.]{\includegraphics[width=.23\textwidth]{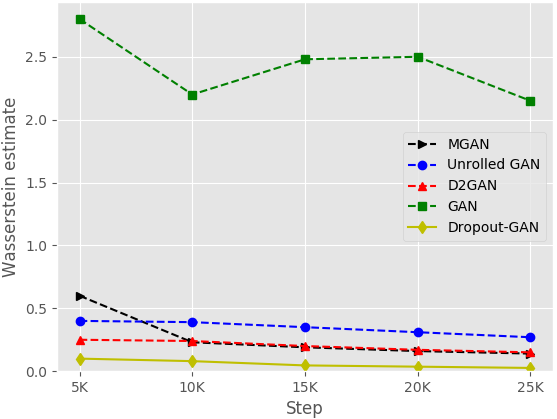}}
   \subfigure[Symmetric KL divergence.]{\includegraphics[width=.23\textwidth]{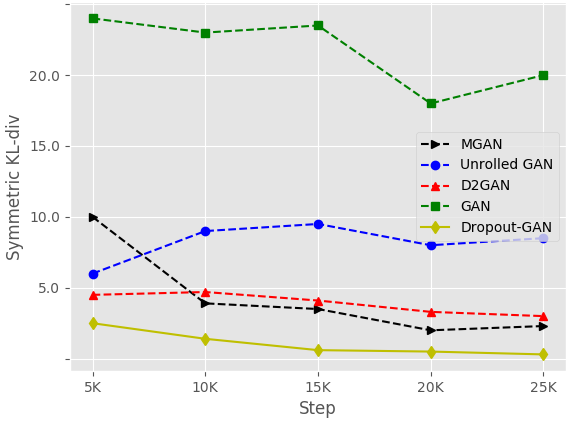}}
\caption{Comparison of Dropout-GAN (8 discriminators) with original GANs, Unrolled GAN, D2GAN, and MGAN (a). Real data is presented in red while generated data is presented in blue. Our method manages to cover all modes with significantly less noisy samples that fall outside any real mode when compared to the other discriminator modified methods. Dropout-GAN also achieves the lowest distance (b) and divergence (c) between the real and generated data, continuously converging to 0 over time.}
\label{fig:toy_dataset}
\end{figure}

We now proceed to compare our approach of applying adversarial dropout to standard GAN, \textit{i.e.,} Dropout-GAN, with other existing methods in the literature. We followed the toy experiment with a 2D mixture of 8 Gaussian distributions (representing 8 data modes) firstly presented in UnrolledGAN~\cite{unrolled_gan}, and further adopted by D2GAN~\cite{dual_GANs} and MGAN~\cite{multi_generator_gans}. We used the same architecture as D2GAN for a fair comparison. The results are shown in Figure~\ref{fig:toy_dataset}, where one can see that Dropout-GAN successfully covers the 8 modes from the real data while having significantly less noisy samples compared to the other discriminator-driven methods. Note that MGAN takes advantage of a multi-generator framework plus an additional classifier network while making use of a different architectural setting. However, due to the simplicity of our approach, we manage to converge to the real data modes faster than the other approach, specifically MGAN, as seen in the early training steps. Moreover, our framework achieves the lowest distance and divergence measures between the real and fake data.


To evaluate the extensibility of our approach, we studied the effects of using adversarial dropout in the following GAN methods: LSGAN, DRAGAN, and standard GAN using both the original (GAN) and modified objective (modGAN). These methods consist of a subset of the methods compared in \cite{are_all_created_gans_equal}, since they cover important variations of the original GAN framework where $D$'s output is either a probability (GAN, modGAN, and DRAGAN) or unbounded (LSGAN), while making use of gradient penalty (DRAGAN) or not. We also followed their presented training settings, training models on MNIST, CIFAR-10, and CelebA for 20, 40, and 100 epochs, respectively. We made use of a simpler architectural setting though, similar to the one previously described in Section \ref{results} but with double the number of neurons per convolutional layer.

\begin{table}
\caption{FID comparisons on MNIST, CIFAR-10 and CelebA. We used 2 discriminators and $d = 0.5$ for all the adversarial dropout methods represented in bold.} 
\begin{center}
\begin{small}
\begin{sc}
\centering
\begin{adjustbox}{width=0.47\textwidth}
\begin{tabular}{lcccr}
\toprule
 & MNIST & CIFAR-10 & CelebA\\
\midrule
Real data    & $\approx$ 0.00 & $\approx$ 0.00 & $\approx$ 0.00\\
\midrule
GAN~(\cite{GANs})  & 22.65 $\pm$ 0.13 & 70.23  $\pm$ 0.07 & 46.18  $\pm$ 0.07\\
\textbf{Dropout-GAN} & \textbf{14.63  $\pm$ 0.18} & \textbf{66.82  $\pm$ 0.10} & \textbf{31.25  $\pm$ 0.09} \\
\midrule
modGAN~(\cite{GANs}) & 22.66  $\pm$ 0.11& 79.58  $\pm$ 0.11 & 41.25  $\pm$ 0.03 \\
\textbf{Dropout-modGAN} & \textbf{15.39  $\pm$ 0.15} & \textbf{67.57  $\pm$ 0.14} & \textbf{35.32 $\pm$ 0.06} \\
\midrule
LSGAN~(\cite{mao2017least})  & 24.05  $\pm$ 0.15 & 83.66  $\pm$ 0.08 & 43.13  $\pm$ 0.04\\
\textbf{Dropout-LSGAN}  & \textbf{15.41  $\pm$ 0.21} & \textbf{69.37  $\pm$ 0.11} & \textbf{37.58  $\pm$ 0.10} \\
\midrule
DRAGAN~(\cite{kodali2017convergence})  & 22.84 $\pm$ 0.15 & 80.57  $\pm$ 0.06 & 46.82  $\pm$ 0.06\\
\textbf{Dropout-DRAGAN} & \textbf{15.20  $\pm$ 0.16} & \textbf{66.90  $\pm$ 0.09} & \textbf{37.21  $\pm$ 0.08} \\
\bottomrule
\end{tabular}
\end{adjustbox}
\end{sc}
\end{small}
\end{center}
\label{fid_comparision}
\end{table}

The best FID scores for each the original and multiple adversarial versions are reported in Table \ref{fid_comparision}. The advantage of using adversarial dropout is significantly visible for each method, lowering the minimum FID obtained considerably for all the tested datasets. For a fair comparison, we used only 2 discriminators when applying adversarial dropout, which makes the overall framework still relatively small with a great benefit in the end results. When simply using an ensemble of discriminators on CIFAR-10, \textit{i.e.} $d = 0$, the proposed dropout variants improve FID by $7.25$, $4.39$, $3.96$, and $9.93$, on GAN, modGAN, LSGAN, and DRAGAN, respectively.

To also test how adversarial dropout behaves on larger datasets, we calculated the Inception Score~\cite{is} (IS) to compare the quality in the same set of methods. On top of CIFAR-10, we further used STL-10~\cite{coates2011analysis}, and ImageNet~\cite{russakovsky2015imagenet}, with the latter two being larger datasets with 100K and 1M images, respectively. We downsized all images to 32x32. We used the same architectures mentioned above. However, we trained each model longer, more specifically 250 epochs for CIFAR-10 and STL-10, and 50 epochs for ImageNet. 

The obtained IS are presented in Table~\ref{is_comparision}. Once again, we observe that applying adversarial dropout considerably increases the obtained IS for all tested datasets, without much overhead since only 2 discriminators were used.

\begin{table}
\caption{Inception score comparisons on CIFAR-10, STL-10, and ImageNet. We used 2 discriminators and $d = 0.5$ for all the adversarial dropout methods represented in bold.} 
\begin{center}
\begin{small}
\begin{sc}
\centering
\begin{adjustbox}{width=0.47\textwidth}
\begin{tabular}{lcccr}
\toprule
 & CIFAR-10 & STL-10 & ImageNet\\
\midrule
Real data  & 11.24  $\pm$ 0.16 & 26.08 $\pm$ 0.26 & 25.78 $\pm$ 0.47\\
\midrule
GAN~(\cite{GANs}) & 5.35 $\pm$ 0.04 & 5.53 $\pm$ 0.03 & 7.30 $\pm$ 0.08 \\
\textbf{Dropout-GAN} & \textbf{6.22 $\pm$ 0.09} & \textbf{7.20 $\pm$ 0.11} & \textbf{7.52 $\pm$ 0.13} \\
\midrule
modGAN~(\cite{GANs}) & 5.49  $\pm$ 0.07 & 6.64 $\pm$ 0.05 & 6.96 $\pm$ 0.08\\
\textbf{Dropout-modGAN} & \textbf{5.90  $\pm$ 0.08} & \textbf{6.95 $\pm$ 0.09} & \textbf{7.26 $\pm$ 0.12}\\
\midrule
LSGAN~(\cite{mao2017least}) & 5.76  $\pm$ 0.05 & 5.32 $\pm$ 0.06 & 6.92  $\pm$ 0.04\\
\textbf{Dropout-LSGAN}  & \textbf{5.95  $\pm$ 0.07} & \textbf{6.88 $\pm$ 0.13} & \textbf{7.08  $\pm$ 0.13}\\
\midrule
DRAGAN~(\cite{kodali2017convergence}) & 5.65  $\pm$ 0.08 & 6.97 $\pm$ 0.09 & 7.41 $\pm$ 0.11 \\
\textbf{Dropout-DRAGAN} & \textbf{6.22 $\pm$ 0.08} & \textbf{7.30 $\pm$ 0.13} & \textbf{7.54 $\pm$ 0.12} \\
\bottomrule
\end{tabular}
\end{adjustbox}
\end{sc}
\end{small}
\end{center}
\label{is_comparision}
\end{table}

\begin{figure}[h]
\centering
\includegraphics[width=0.47\textwidth]{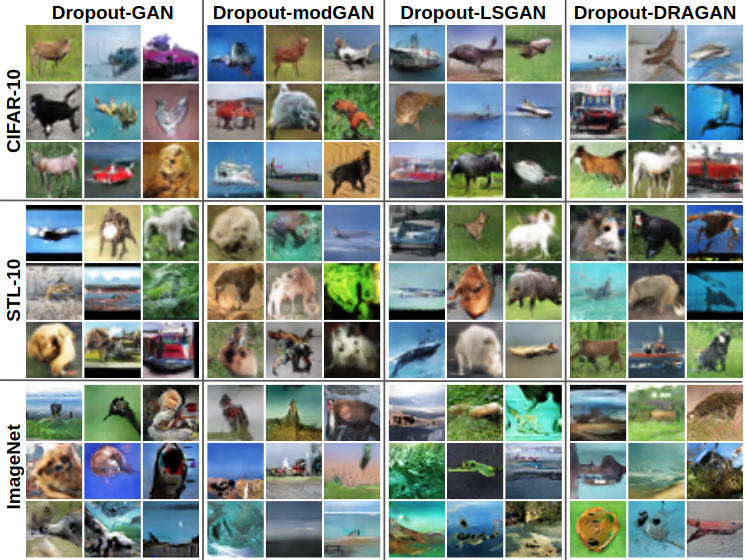}
\caption{CIFAR-10, STL-10, and ImageNet generated samples for the different GAN approaches using adversarial dropout.}
\label{fig:cifar_stl_imagenet}
\end{figure}

A subset of randomly generated samples for each method when using adversarial dropout is presented in Figure~\ref{fig:cifar_stl_imagenet}, where one can see high diversity alongside with high quality, even on the bigger datasets. These results solidify the success of mitigating mode collapse when applying the adversarial dropout to the different methods.

Finally, we directly compare the quality of the generated samples between Dropout-GAN, GMAN~\cite{gman}, and original GANs~\cite{GANs} with the modified loss, using both 2 and 5 discriminators on CIFAR-10. In their original experiments, \cite{gman} used Inception Score~\cite{is} as the evaluation metric, with higher values correlating to better generated samples. For a fair direct comparison, we used the same architectures and training procedures as originally used in GMAN's experiments. Results are presented in Table~\ref{gman_comparision}. Dropout-GAN outperforms both methods for all different number of discriminators scenarios on all tested dropout rates.

\begin{table}
\caption{Inception score comparison between Dropout-GAN and different variants of GMAN on CIFAR-10. Original GANs with the modified loss is also presented as a baseline.}
\label{gman_comparision}
\begin{center}
\begin{small}
\begin{sc}
\begin{adjustbox}{width=0.47\textwidth}
\begin{tabular}{lcccr}
\toprule
 & 1 disc. & 2 disc. & 5 disc.\\
\midrule
GAN~(\cite{GANs})   &  5.74 $\pm$ 0.17& - & -\\
\midrule
GMAN-0~(\cite{gman}) & - & \underline{5.88 $\pm$ 0.19} & 5.96 $\pm$ 0.14\\
GMAN-1~(\cite{gman}) & - & 5.77 $\pm$ 0.16 & \underline{6.00 $\pm$ 0.19}\\
GMAN*~(\cite{gman}) & - & 5.54 $\pm$ 0.09 & 5.96 $\pm$ 0.15\\
\midrule
Dropout-GAN $(d=0.2)$ & - & 5.95 $\pm$ 0.10 & 6.01 $\pm$ 0.12\\
Dropout-GAN $(d=0.5)$ & - & \underline{\textbf{5.98 $\pm$ 0.10}} & \underline{\textbf{6.05 $\pm$ 0.15}}\\
\bottomrule
\end{tabular}
\end{adjustbox}
\end{sc}
\end{small}
\end{center}
\end{table}

\section{Related Work}
\label{related_work}




We will now focus on previous work that mitigated mode collapse in GAN. Instead of extending the original framework to multiple adversaries, one can change GAN objective to directly promote sample diversity. WGAN~\cite{wgan} and MMD GAN~\cite{mmd_gan} proposed to optimize distance measurements to stabilize training. On the other hand, EBGAN~\cite{energy_gan} and Coulomb GANs~\cite{coulomb_gan} reformulated the original GAN problem using an energy-based objective to promote sample variability. While Regularized-GAN and MDGAN~\cite{mode_regularized_gans} make use of an autoencoder to penalize missing modes and regularize GAN objective, DFM~\cite{warde2016improving} makes use of autoencoders to perform high-level feature matching. UnrolledGAN~\cite{unrolled_gan} changes $G$ objective to satisfy an unrolled optimization of $D$. LSGAN \cite{mao2017least} proposes to use a least-squares loss for the $D$ while DRAGAN \cite{kodali2017convergence} applies gradient norm penalty on top of original GAN.

Although some work has focused on augmenting the number of generators \cite{GhoshKN16,GhoshKNTD17,multi_generator_gans}, or even increasing both the number of generators and discriminators \cite{boosted_generative_models,triangle_gan,Chavdarova_CVPR_2018}, we turn our focus on methods that solely increases the number of discriminators to prevent mode collapse. D2GAN~\cite{dual_GANs} proposed a single generator dual discriminator architecture where one $D$ rewards samples coming from the true data distribution, while the other rewards samples that are likely to come from $G$. Thus, each $D$ still operates on a different objective function. GMAN~\cite{gman} proposed a framework where a single $G$ is trained against several discriminators on different levels of difficulty, by either using the mean loss of all discriminators (GMAN-0), picking only the $D$ with the maximum loss in relation to $G$'s output (GMAN-1), or controlled by $G$ through a hyperparameter $\lambda$ (GMAN*). Recently, microbatchGAN~\cite{mordido2020microbatchgan} assigned a different portion of each minibatch to each discriminator to stimulate sample diversity. 

However, all of the described approaches have some sort of constraints, either by restricting each $D$'s architecture to be different, or by using different objective functions for each $D$. We argue that these are limitations from an extensibility point of view, none of which exists in our proposed framework. Moreover, we note that applying Dropout-GAN's principles of using adversarial dropout to the previously described methods would be a viable step to further promote sample diversity.

\section{Conclusion and Future Work}
\label{conclusion}

In this work, we propose to mitigate mode collapse by proposing a new framework, called Dropout-GAN, that enables a single generator to learn from an ensemble of discriminators that dynamically changes at the end of every batch by use of adversarial dropout. We conducted experiments on multiple datasets of different sizes that show that adversarial dropout successfully contributes to a bigger sample variety on multiple GAN approaches. Moreover, it also increases training stability over time by enabling $G$ to receive more quantity and variety of feedback.

In the future, it would be interesting to adjust $G$'s learning rate according to the size of the discriminator set, allowing a more coherent learning speed between $G$ and each $D$, especially when using a large ensemble.
Moreover, applying game theory to make the different discriminators dependent, \textit{i.e.,} aware of each other's feedback, could also be a very interesting path to follow, taking full advantage of using multiple adversarial training.

\bibliography{ecai}
\end{document}